\documentclass[letterpaper, 10 pt, conference]{ieeeconf}  % Comment this line out if you need a4paper

\IEEEoverridecommandlockouts                              % This command is only needed if 
\usepackage{tikz}
\usetikzlibrary{positioning, fit, calc, chains, matrix}

\usepackage[ruled,vlined]{algorithm2e}
\usepackage[pagebackref=true,breaklinks=true,letterpaper=true,colorlinks,bookmarks=false]{hyperref}

\title{\LARGE \bf
Bridging Intelligence and Instinct: A New Control Paradigm for Autonomous Robots
}

\author{Shimian Zhang$^{1}$ % <-this % stops a space
\thanks{$^{1}$Shimian Zhang is with Shanghai Hrstek Co., Ltd
        {\tt\small zhangsm@hrstek-robotics.com}}%
Qiuhong Lu$^{2}$ % <-this % stops a space
\thanks{$^{2}$Qiuhong Lu is with Shanghai Hrstek Co., Ltd
{\tt\small luqh@hrstek-robotics.com}}%
}

\begin{document}

\maketitle
\thispagestyle{empty}
\pagestyle{empty}

%%%%%%%%%%%%%%%%%%%%%%%%%%%%%%%%%%%%%%%%%%%%%%%%%%%%%%%%%%%%%%%%%%%%%%%%%%%%%%%%
\begin{abstract}
As the advent of artificial general intelligence (AGI) progresses at a breathtaking pace, the application of large language models (LLMs) as AI Agents in robotics remains in its nascent stage. A significant concern that hampers the seamless integration of these AI Agents into robotics is the unpredictability of the content they generate, a phenomena known as ``hallucination''.
Drawing inspiration from biological neural systems, we propose a novel, layered architecture for autonomous robotics, bridging AI agent intelligence and robot instinct. 
In this context, we define Robot Instinct as the innate or learned set of responses and priorities in an autonomous robotic system that ensures survival-essential tasks, such as safety assurance and obstacle avoidance, are carried out in a timely and effective manner. 
This paradigm harmoniously combines the intelligence of LLMs with the instinct of robotic behaviors, contributing to a more safe and versatile autonomous robotic system. As a case study, we illustrate this paradigm within the context of a mobile robot, demonstrating its potential to significantly enhance autonomous robotics and enabling a future where robots can operate independently and safely across diverse environments.

\end{abstract}

%%%%%%%%%%%%%%%%%%%%%%%%%%%%%%%%%%%%%%%%%%%%%%%%%%%%%%%%%%%%%%%%%%%%%%%%%%%%%%%%
\section{Introduction}

The rapid evolution of artificial general intelligence (AGI) technologies, particularly large language models (LLMs) like GPT-4 \cite{openai2023gpt4} and LLaMA \cite{touvron2023llama}, has catalyzed a new wave of potential applications within the realm of robotics \cite{vemprala2023chatgpt}. As AI agents, LLMs can generate high-level decisions and instructions that guide a robot's behavior , demonstrating potential in tool utilization \cite{liang2023taskmatrix, shen2023hugginggpt}, task planning \cite{nakajima2023task},and task creation \cite{bubeck2023sparks}. However, despite the promises, there remains a dearth of substantive applications within robotics, largely due to the unpredictability of LLMs' outputs, a phenomenon often referred to as ``hallucination''  \cite{openai2023gpt4, bang2023multitask}.

\begin{figure}[t]
\centering
\includegraphics[width=\linewidth]{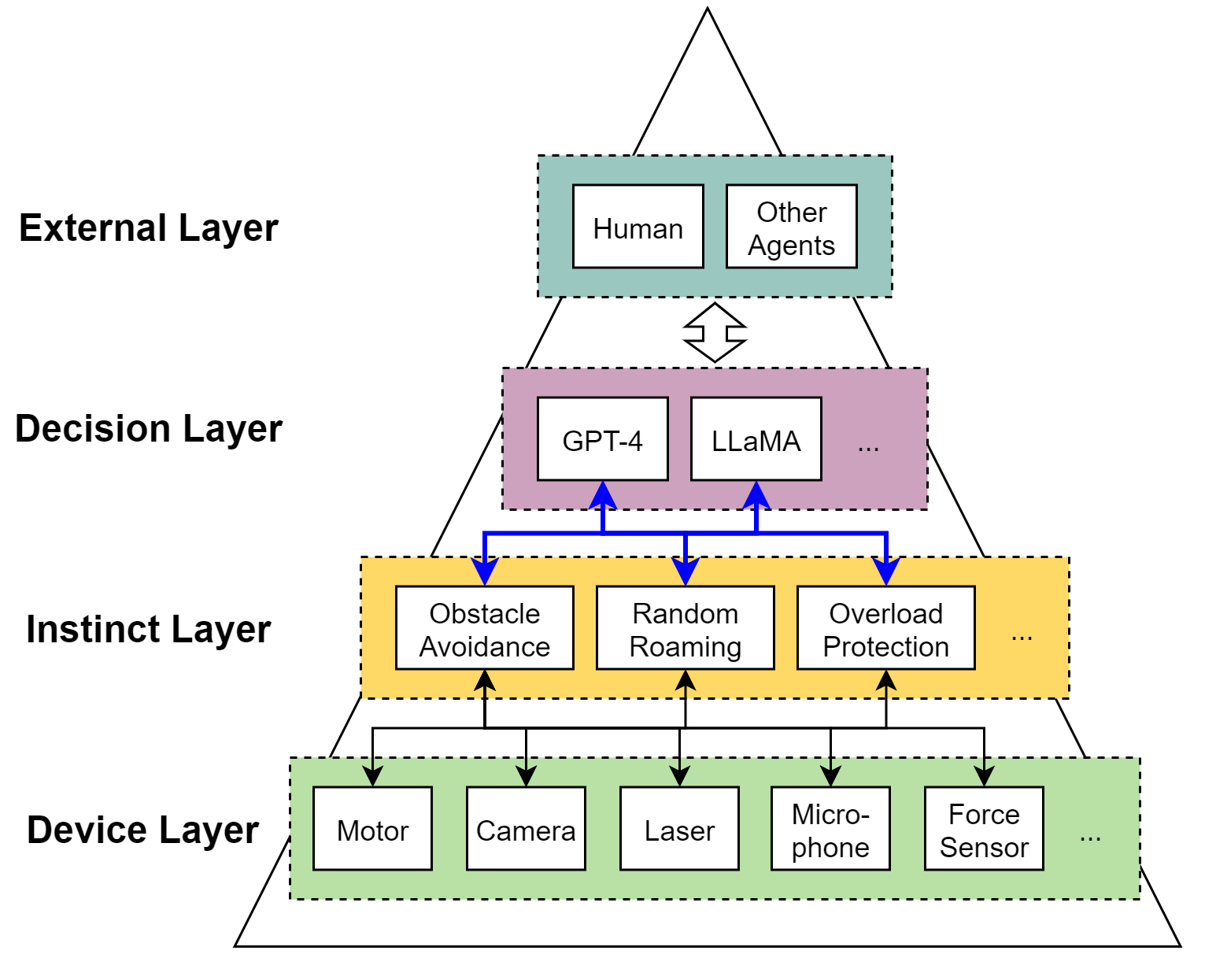}
\caption{Layered Hierarchy Design. The pyramid structure of our proposed architecture, comprising four layers: External, Decision, Instinct, and Device. The External layer represents high-level entities interacting with the system, the Decision layer acts as the `brain' making high-level decisions, the Instinct layer continuously maintains safety acting as the `brainstem,' and the Device layer executes the commands by controlling the robot's physical actions.}
\label{fig:framework}
\end{figure}

Mitigation strategies such as Chain-of-Thoughts \cite{kojima2022large}, Self-Reflection \cite{shinn2023reflexion} and etc., have been proposed to manage this issue, but these interventions target the LLMs themselves rather than considering potential architectural solutions within the robotic system. Traditionally, robotic control systems have adhered to either cognitive models \cite{Levesque_Lakemeyer_2008}, which mimic human cognitive processes, or behavior-based models that produce direct responses to sensory inputs \cite{brooks1986robust}. Yet, none of these traditional architectures have fully considered the incorporation and interaction of AI agents.

In response to these challenges, we introduce a novel architecture for robotic control systems designed to bridge high-level intelligence with low-level instinctual protocols, as Fig.~\ref{fig:framework} shows. Inspired by the human nervous system's 'brain and brainstem' paradigm, this architecture proposes four distinct layers: External, Decision, Instinct, and Device. This new paradigm treats the AI agent as the 'brain,' handling advanced decision-making, while a 'Robot Instinct' module acts as the 'brainstem,' overseeing basic survival-essential tasks.

Our architectural design presents a systemic solution to the hallucination risks posed by LLM-based AI agents. By integrating high-level decision-making AGIs with robust low-level safety mechanisms, we limit the harm of potential incorrect decisions. Moreover, we emphasize the need for robots, even as they gain sophisticated AGI capabilities, to retain robust instinctual reactions akin to human survival instincts, ensuring they can effectively serve in various tasks and environments.

The primary contribution of this paper is the introduction of this innovative four-layered architecture for robotic control systems. We demonstrate its efficacy and versatility through a specific case study, marking a significant stride towards a new era of robotics. We envision this approach not only propelling the development of robotic control systems but also fostering a tighter synergy between AI and robotics, empowering more nuanced interaction between robots and their environment.

% This paper is organized as follows: Section 2 presents related works and positions our study within the context of the existing body of knowledge. Section 3 details our proposed architectural framework. Section 4 provides a case study illustrating the application of our framework within a mobile robot context. Finally, Section 5 concludes the paper and suggests potential directions for future work.

\section{Related Works}
\subsection{AI Agents for Robotics}

The application of AI agents in robotics has become a burgeoning field of research with the advancement of machine learning techniques, especially Large Language Models (LLMs).

\subsubsection{LLMs as AI Agents}
Recent research findings indicate that LLMs can be effectively used as AI agents to address complex tasks in robotics \cite{bubeck2023sparks, liang2023taskmatrix, shen2023hugginggpt, nakajima2023task}. Such capabilities enable LLMs to interact with both structured and unstructured environments, making them valuable for the development of versatile and adaptable robotic systems.

\subsubsection{AI Agents as the Robot Brain}
Despite these advancements, the application of AI agents as the primary "brain" of a robot is still in the early stages \cite{mai2023llm}, largely limited by the complexities of integrating AI with physical systems. Some work has integrated models like Chat-GPT into a robot's operational flow \cite{vemprala2023chatgpt}, albeit in a simulated environment with the need for human oversight to ensure accuracy and appropriateness of the AI's outputs.

In light of these challenges and opportunities, our proposed design framework represents a pioneering step towards fully integrating AI agents into robotic systems. We view this as an inevitable trend in the evolution of robotics, given the increasing demand for intelligent and autonomous operation in real-world environments. Our approach aims to create a truly anthropomorphic robotic entity, capable of sophisticated interaction with the world while ensuring the safety and reliability of operations.
% The proposed design leverages the power of AI while addressing some of the inherent challenges, marking a significant step forward in the field of robotics.

% \subsection{Robotic Control Systems}
\subsection{Robot System Architectures}

Historically, the Sense-Plan-Act (SPA) architecture was proposed as a tiered model with distinct modules \cite{saridis1983intelligent}, each responsible for a specific function or task. While this top-down architecture was effective in dividing labor, it was also inherently sequential, resulting in difficulties in rapidly adapting to dynamic environments.

An alternative approach was put forth by \cite{brooks1986robust}, who proposed the Subsumption Architecture. This model popularized Behavior-Based Robotics (BBR), which emphasized decentralized control and modularity. Complex behaviors were seen as emergent from the interaction of simple, concurrently operating modules. However, the practical implementation of this architecture has proven to be complex, particularly in intelligent robots where the decoupling of high-level and low-level behaviors can be challenging.

The Hybrid Architecture was then introduced to bridge the strengths of the SPA and BBR architectures. In a hybrid model, a robot's behavior can be generated either by a central plan (akin to SPA) or by a set of independent behavior modules (akin to BBR). \cite{gat1998three} proposed a notable instance of this paradigm, a three-tiered architecture consisting of a reactive Controller layer, a reactive Sequencer layer, and a Deliberator layer.

Our proposed paradigm control architecture could be seen as a new form of Hybrid Architecture, introducing unique components designed to maximize the benefits of recent advancements in AI. It innovatively blends the deliberative power of AI Agent decision-making capabilities with the reactivity of an Instinct layer witth low-level safety mechanisms, coupled with a clear definition of human-machine interactions within the External Layer.

\subsection{Safety Mechanisms}
% The critical task of ensuring safety in robotic systems has been approached through various methodologies, including safety policies, post-impact safety protocols, and control barrier functions. Safety policies, often backed by reinforcement learning, incentivize robots to achieve tasks while preserving safety. Task-Based Post-Impact Safety Protocols handle unexpected collisions, prompting appropriate reactions to minimize potential damage. Control barrier functions offer mathematical guarantees of safety by constraining the control input, ensuring the system operates within safe bounds. In our framework, the Robot Instinct layer integrates these safety mechanisms, ensuring robust protection independently of the AI Agent's decision-making layer.

Safety mechanisms are crucial to the design and implementation of any robotic system, ensuring safe interactions with the environment and reliable execution of tasks. Various techniques have been developed over the years. 

\subsubsection{Model-Based Safety Mechanisms}
Techniques such as Control Lyapunov Function (CLF) and Control Barrier Function (CBF) have been extensively used to enforce safety in robotic systems \cite{ames2019control, nguyen2015safety, gong2019feedback}. These techniques are highly effective in systems where precise mathematical models of the system dynamics and environment are known. However, the performance of these methods can be significantly degraded if the model is inaccurate, or in the presence of unknown system dynamics.

\subsubsection{Model-Free Safety Control}
With the advent of powerful Reinforcement Learning (RL) algorithms, model-free safety control mechanisms have emerged as promising solutions on complex and high-dimensional systems \cite{hwangbo2017control, li2021reinforcement}. These mechanisms are particularly useful when dealing with unknown environments or unmodeled system dynamics. However, RL-based mechanisms typically require a large amount of sample data for training and can be time-consuming. Such mechanism are usually not easy to migrate from one robot platform to another.

\subsubsection{AI Agents Safety Concerns}
Given the rapidly growing capabilities of LLMs, it is imperative to address the potential safety concerns that come with their integration into robotic systems. Techniques such as Chain of Thoughts \cite{kojima2022large}, Self-Reflection \cite{shinn2023reflexion}, and human-in-the-loop mechanisms \cite{vemprala2023chatgpt} have been used to prevent or reduce the rate of "hallucination". However, these methods do not entirely eliminate the propensity of LLMs to generate incorrect responses due to their inherent nature.

Our architecture distinctly addresses these safety concerns through two primary mechanisms: First, our  Instinct Layer operating independent, uninterrupted safety mechanisms, effectively caters to safety requirements. The implementation of these safety protocols can leverage established safety control methods such as CBF, RL-learning, etc. Second, our Decision Layer has a multi-tiered interaction mechanism including  a feedback loop with the Instinct Layer for continual self-reflection and interaction with the External Layer for incorporating human-in-the-loop to prevent the possibility of incorrect planning from an LLM as much as possible.

\section{Proposed Framework}
\subsection{Layered Hierarchy Design}

Our proposed control framework revolutionizes the conventional robotic architecture by incorporating a hierarchically layered design based on the level of intelligence, as Fig.~\ref{fig:framework} shows. This hierarchical model emulates the human cognitive process, enabling more human-like behavior and interaction in robotic systems.

\subsubsection{External Layer}
At the top of this hierarchy, as an external layer, we have the (Artificial) General Intelligence, represented by humans and other high-level intelligent agents. This level provides high-level goals and instructions and serves as a means of interpreting the robot's behavior and feedback in the broader context of the world.
\subsubsection{Decision Layer}
Below this, we have the Decision Layer, the highest level within the robot's architecture. This layer is composed of multiple AI Agents as the 'brain' of the robot, each potentially an instance of high-level generative models such as GPT-4, LLaMA, etc. 
The tasks carried out by these AI Agents encompass a range of high-level goals that are fundamental for a fully autonomous robot. Such tasks include complex interaction with humans or other agents, autonomous decision making based on the environment or the tasks at hand, goal-directed behavior based on both short-term and long-term objectives, and even tool use and learning. 

Realizing these high-level tasks is often achieved by converting the tasks into a text form that the AI models understand [citation]. The models can then generate a chain of thoughts or decisions [citation], similar to how humans would reason. This makes use of AI Planners and AI Executors [citation], which respectively deal with deciding what to do and implementing the decision.

\subsubsection{Instinct Layer}
Following the Decision Layer is the Instinct Layer. This level of our framework represents the low-level intelligence within the robotic system as the ‘brainstem’ and is composed of a collection of Robot Instinct modules. Unlike the AI Agents that make strategic decisions, the role of the Robot Instinct modules is to ensure the robot's safety and operability in the face of immediate environmental challenges. Each of these modules can be powered by a Discriminative Model, or a more traditional control scheme depending on the specific task.

These \textbf{survival-essential} tasks are abilities intrinsic to the robot's survival and functionality, such as obstacle avoidance, random roaming, and overload protection. These functions remain operational even in the absence or failure of the AI Agents, whether due to hallucination phenomena, signal interference, or other unexpected scenarios. Consequently, even when higher-level decision-making capabilities are compromised, the robot can still provide basic services and ensure its own safety and that of its surroundings.

\subsubsection{Device Layer}
Finally, at the base of the hierarchy, we have the Device Layer. This layer is devoid of any form of intelligence and comprises the fundamental hardware components of the robot, such as the motor, camera, laser, and DRAM. These devices serve as the robot's sensors, executors, and memory, forming the 'physical body' of the robot that interacts with the environment.

By adopting such a layered, hierarchical design, our framework allows for the distribution of control and decision-making processes at different levels of intelligence. This not only enables robots to exhibit more human-like behavior but also opens new avenues for safer and more reliable robot operation.

\subsection{Modules and Flows}
\begin{figure}
    \centering
    \includegraphics[width=\linewidth]{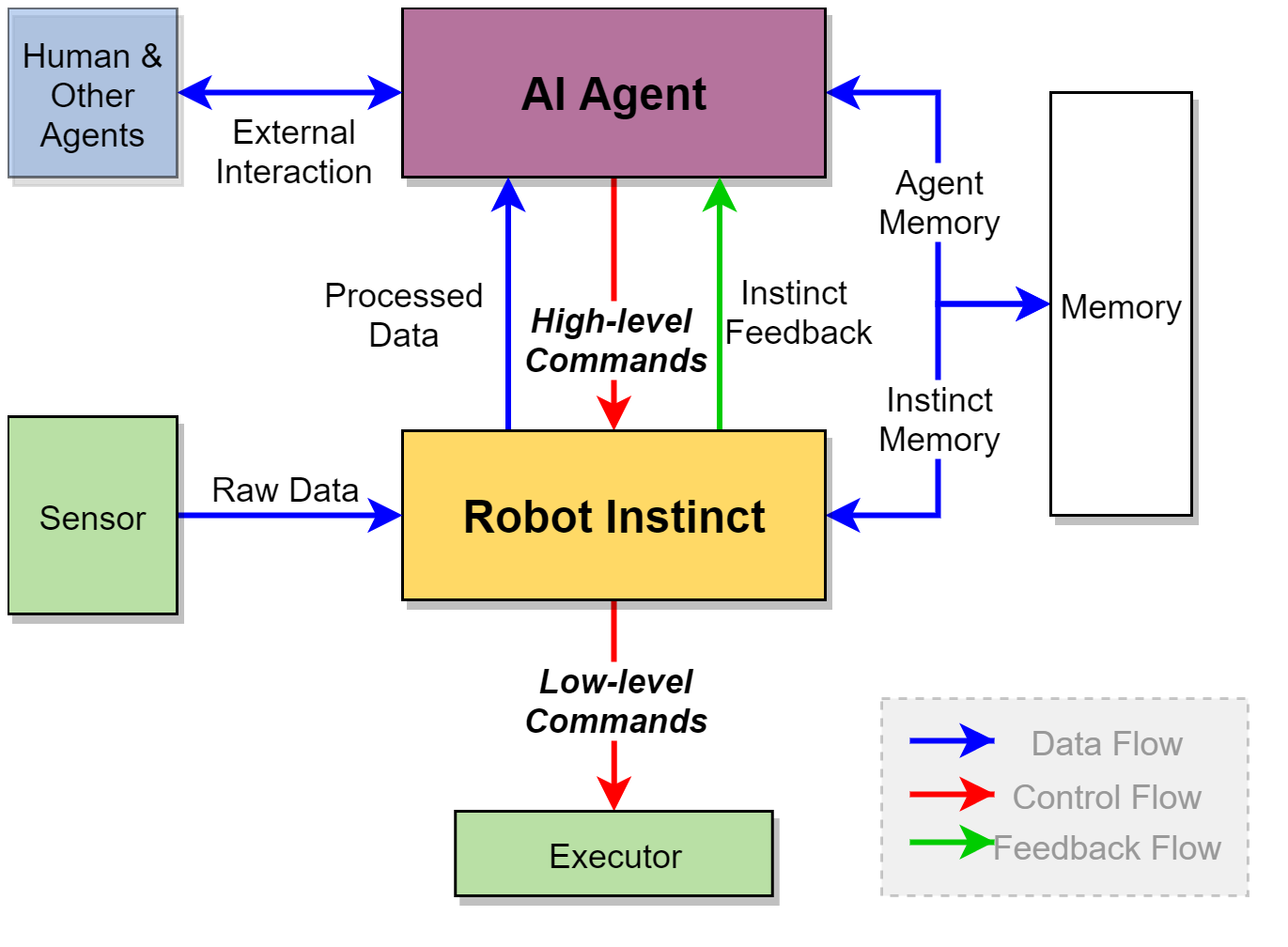}
    \caption{Modules and Flows. 
    Drawing from our layered hierarchy, we can distill our framework into a few key modules: the (Artificial) General Intelligence module, which includes Human and/or Other Agents, the AI Agent module, the Robot Instinct module, and the Device Layer modules. The Device Layer is further subdivided into the Sensor, Executor, and Memory modules. This modular breakdown allows for targeted interactions and efficient data and control flows within the robotic system.
    }
    \label{fig:module_flow_diagram}
\end{figure}

After having delved into the detailed discussion on the layered design of the architecture, we now turn to delineating the functionalities of the various modules within the architecture and the data, control, and feedback flows between them, as Fig.~\ref{fig:module_flow_diagram} shows. This modular design and flow analysis are key to realizing our layered architecture, jointly forming the backbone of the architecture.

\subsubsection{Data Flow}
All real-time data (such as image and audio signals and motor encoder information) are first acquired by the Sensor module and then delivered to the Robot Instinct module. The Robot Instinct is responsible for processing this raw sensory data, executing survival-essential tasks, and producing a simplified version for the AI Agent. This design eases the computational burden of the AI Agent and allows it to focus more on high-level decision-making, similar to how higher cognitive functions in humans rely on processed sensory inputs.

The processed data from Robot Instinct is then transferred to the AI Agent for advanced decision-making processes. Both the Robot Instinct and AI Agent generate data based on their operation, which is stored in the Memory module for future recall and learning. The real-time nature and reliability of the Robot Instinct processing are critical to the successful operation of this framework.

\subsubsection{Control Flow}
The AI Agent, using generative AI, generates high-level commands, often as function API calls in high-level languages, even pseudocode in complex conditions or loops. These commands, which encapsulate complex tasks or behaviors, are then sent to the Robot Instinct module.

The Robot Instinct module is responsible for the execution of these high-level commands. To do so, it translates high-level commands into a series of low-level API calls such as specific motor movements or stops, in languages closer to the hardware, like C \footnote{The Robot Instinct's safety check should ensure low latency, allowing the high-level commands to "penetrate" through the Robot Instinct and reach the Device Layer directly while maintaining safety.}. 

Importantly, the Robot Instinct module provides a standardized interface for the AI Agent, abstracting the details of the underlying Executor devices. This abstraction allows the AI Agent to focus on high-level decision-making without worrying about the specific characteristics of the individual devices, which greatly enhances the scalability and versatility of the AI system.

\subsubsection{Feedback Flow}
Feedback data from the Robot Instinct module is returned to the AI Agent, promoting a process of self-reflection and adjustment. This feedback mechanism allows the AI Agent to make informed next-step decisions, thus enabling more effective learning and adaptation to dynamic environments. 

\subsection{Feedback Mechanism}
\begin{figure}
    \centering
    \includegraphics[width=\linewidth]{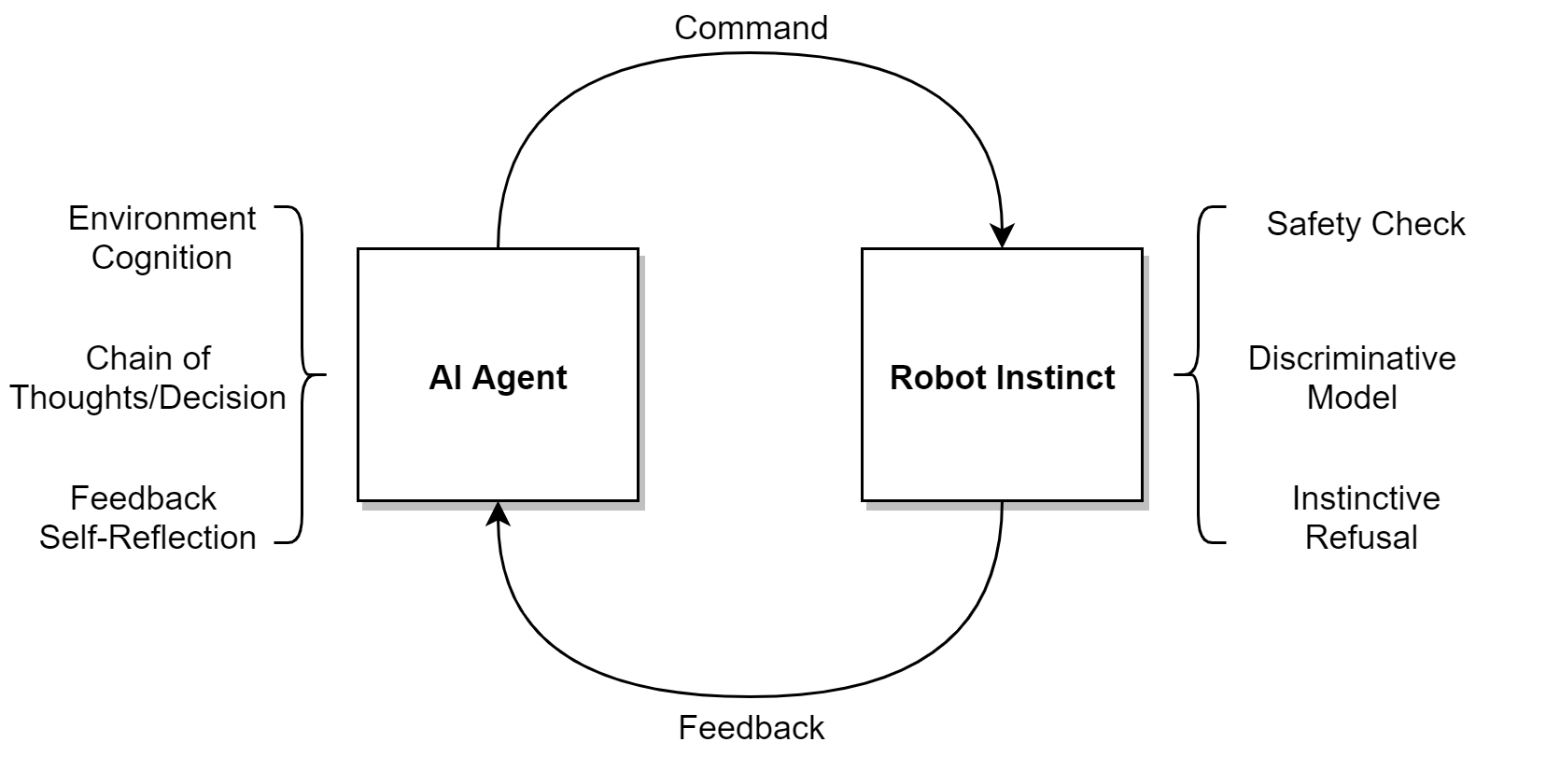}
    \caption{Feedback Mechanism.}
    \label{fig:feedback_diagram}
\end{figure}

In our proposed robotic control framework, the AI Agent and the Robot Instinct interact with each other not merely through one-way command issuing but also through bidirectional feedback communication, as Fig.~\ref{fig:feedback_diagram} shows. This closed-loop structure aids in refining the decision-making process and ensures the safety of the robot's operations.

\subsubsection{Feedback to the AI Agent}
The AI Agent, after issuing high-level commands, receives feedback from the Robot Instinct. This feedback includes the status of the commands executed and possibly additional sensor data. By incorporating modern techniques such as chain of thoughts/decision, self-reflection via in-context learning, the AI Agent can optimize subsequent commands without the need for retraining the model. This is a critical feature for the robot to learn and adapt to new environments and situations.

\subsubsection{Safety Protocols and Instinctive Refusal}
On the other side, the Robot Instinct module is not simply a recipient of high-level commands from the AI Agent. It has built-in safety protocols that allow it to check every incoming command. The safety check utilizes either a discriminative model or traditional methods to ensure the commands will not violate any \textbf{survival-essential} tasks. If a command is deemed unsafe, the Robot Instinct module has the authority to instinctively refuse it. This feature ensures that the robot’s fundamental safety and self-preservation are always a priority.

\subsection{Architecture Revisit: Intelligence and Necessity}
We now revisit our layered design from dual perspectives. We delve further into the framework’s intelligence and necessity, providing a comprehensive understanding and evaluation of our design.

Firstly, our layered architecture presents a top-down distribution in terms of the level of intelligence. At the highest level, in the External Layer, we find humans or other advanced intelligent agents capable of sophisticated decision-making and high-level task planning. The Decision Layer houses the AI Agent, responsible for the robot's high-level decisions. Following is the Instinct Layer, accountable for some critical survival tasks such as obstacle avoidance and overload protection. Finally, at the base in the Device Layer, we have basic devices and systems such as sensors and actuators.

Secondly, our layered architecture demonstrates a bottom-up distribution in terms of necessity. At the base, the Device Layer is the necessary structural part of the robot, without which the robot cannot achieve basic perception and action. The Instinct Layer is the necessary control part of the robot, without which the robot is unsafe and incapable of performing basic survival tasks. The Decision Layer is a necessary part of the intelligent robot, without which the robot cannot perform advanced decision-making and task planning. Finally, at the highest level, the External Layer is a necessary part for group robots or human-robot interaction, without which the robot cannot perform cooperation and interaction.

\section{Case Study: Mobile Robot}

\subsection{The Traditional Behavior-based Structure}

In this section, we examine the traditional behavior-based structure by means of a mobile robot case study. This existing control framework is structured around distinct parallel behavior modules, each with their direct input from sensors and direct output to actuators. These behavior modules, acting like individual 'mini-brains,' coordinate to control the robot. Key behavioral modules include target tracking, map building (e.g., based on LiDAR SLAM), path planning, motor operation, and obstacle avoidance modules.
% , as Fig. shows.

The behavior-based structure has its advantages. It stands out for its simplicity, which leads to efficiency in design and testing stages. This design approach also ensures that individual modules work independently, minimizing mutual interference and allowing for independent optimization of each.

However, the main drawback of the behavior-based structure emerges in the context of system generalization and scalability. In this architecture, every behavioral module receives its input directly from sensors and sends its output directly to actuators. As a result, each module needs to be intimately familiar with the particular sensors and actuators of the robot it is implemented on. This specificity restricts the reusability of these modules across different robotic platforms and types, hindering the potential for broad-based solutions and extensions to various robotic platforms. Moreover, due to the tight coupling between modules and specific hardware, making adjustments for new sensors or actuators, or adding new behaviors, may require significant system rewrite, adding complexity to the system's maintenance and expansion.

\subsection{Transition to the Proposed Framework}
Transiting from traditional behavior-based structures, we elucidate our novel framework from bottom to top.

\subsubsection{Device Layer}
The Device Layer comprises the fundamental hardware necessary for the operation of the robot, including motors, actuators, sensors, and other physical components that directly interact with the environment.

\subsubsection{Instinct Layer}
% This layer consists of the Robot Instinct module. Here, critical functions such as obstacle avoidance and motor control are handled autonomously, akin to our human instincts. These operations are made safe and efficient by using traditional control algorithms or decision-based models. The instinct layer primarily interacts with the device layer to acquire sensor data and send actuator commands. It also communicates with the Decision Layer by processing raw data and providing it for further decision making, as well as receiving high-level commands.

The Instinct Layer acts as a critical hub, orchestrating survival-essential tasks such as obstacle avoidance, overload protection, and safety assurance to prevent harm to humans. Different modules within this layer, each running on a dedicated chip, handle these tasks in parallel, achieving both efficiency and robustness. This design enables the layer to adapt to escalating complexities as mobile robot tasks increase in their scope and intricacy.

To ensure the robust and timely execution of these tasks, the Instinct modules are implemented on low-latency, high-stability platforms such as micro-controllers or FPGAs. As depicted in Alg.~\ref{alg:instinct}, each instinct module has the highest thread priority. This priority assignment ensures that survival-essential tasks are always addressed promptly, even in the face of concurrent high-level command executions from the Decision Layer.

A significant feature of our architecture is the dual-layer safety mechanism incorporated into the Instinct Layer. Every high-level command received from the Decision Layer must pass a rigorous safety check before execution. This procedure guarantees that any command, whether survival-essential or derived from higher-level decision-making, conforms to the pre-defined safety protocol. Thus, our architecture harmonizes advanced decision-making capabilities with rigorous safety assurances, providing a reliable and safe control system for mobile robots.

% The Instinct Layer is responsible for executing survival-essential tasks such as obstacle avoidance, overload protection, and safety assurance to prevent harm to humans. As the complexity of mobile robot tasks increases, the scope of these survival-essential tasks also expands. Algorithm~\ref{alg:instinct} outlines a pseudo-algorithm for the Instinct module. This module is implemented on a low-latency, high-stability microcontroller or FPGA to ensure robust operation. It is assigned the highest thread priority, ensuring that survival-essential tasks are performed before executing high-level commands from the Decision Layer. A dual-layer safety mechanism is incorporated, with a safety check performed for both survival-essential tasks and high-level commands.

\begin{algorithm}
\SetAlgoLined
\KwIn{High-level commands from Agent, Devices (Motor, Lidar)}
\KwOut{Low-level command execution, Feedback to Agent}
initialization\;
\While{True}{
    \tcp{Survival Essential Tasks}
    \If{device.status == "safe"} {
        performSurvivalTasks()\;
    } \Else {
        enterSafeMode()\;
        sendFeedback()\;
        continue\;
    }
    
    \tcp{Handling High-level Commands}
    commands = getHighCommands()\;
    \ForEach{command}{
        lowCommands = convert(command)\;

        \ForEach{lowCommand}{
            \If{safetyCheck(lowCommand)}{
                \Switch{lowCommand.type}{
                \Case{Motor}{
                    motor.execute(lowCommand)\;
                }
                \Case{Lidar}{
                    lidar.acquire(lowCommand)\;
                }
        	    }
                sendFeedback()\;
                sendData()\;
            }\Else{
                refusal()\;
                sendFeedback()\;
            }
        }
    }
}
\caption{Instinct Module}
\label{alg:instinct}
\end{algorithm}

\subsubsection{Decision Layer}

The Decision Layer serves as the 'brain' of the robot, where intricate decision-making tasks are carried out. As highlighted in Algorithm~\ref{alg:agent}, this layer is not operated by a single LLM, but rather, it consists of multiple LLM modules that collaborate to perform the AI Agent's tasks. These tasks include interactions with humans or other agents (facilitated by the Interaction Agent), task planning and order arbitration (handled by the Task Planning Agent), tool utilization (enabled by the Tool Utilization Agent), and self-reflection based on feedback from the Instinct Layer (performed by the Self-Reflection Agent).

The Interaction Agent enables the AI to comprehend and respond to external instructions, facilitating cooperative tasks between the robot and humans or other agents. The Task Planning Agent, working closely with the Self-Reflection Agent, enables the AI to analyze dynamic environments and make autonomous decisions. By interpreting the Instinct Layer's API calls/documentation, the Tool Utilization Agent enhances the system's versatility.

Incorporated within the Decision Layer is a safety mechanism that reduces the risk of incorrect AI agent decisions. By assigning distinct roles to each LLM module and establishing feedback loops, the AI is less prone to making mistakes. The safety checks at the Decision Layer further reinforce the safety precautions implemented at the Instinct Layer, providing a comprehensive safety system for the robot.

\begin{algorithm}
\SetAlgoLined
\KwIn{Tasks from External Layer, Feedback from Instinct Layer}
\KwOut{High-level commands to Instinct Layer}
initialization\;
\While{True}{
    \tcp{LLM for interaction with External Layer}
    tasks = getTasks()\; 
    \ForEach{task}{
        \While{task not complete}{
            \tcp{LLM for getting feedback from Instinct Layer}
            feedback = getFeedback()\;
            status = getData()\;
            \tcp{LLM for self-reflection}
            reflect = selfReflection(feedback, status)\;
            \tcp{LLM for task planning and order arbitration}
            highCommands = plan(task, reflect, status)\;
            \tcp{LLM for tool utilization}
            sendCommand(highCommands)\;
            task.update()\;
        }
    }
}
\caption{Agent Algorithm with Multiple LLMs}
\label{alg:agent}
\end{algorithm}

\subsubsection{External Layer}
The External Layer represents high-level (artificial) general intelligence, including humans and other AI agents that send out commands to the robot. In an era demanding advanced human-machine cooperation and swarm coordination, our External Layer is designed to meet these needs by facilitating complex and sophisticated interactions and commands.

% This layer facilitates interaction with other robots, human operators, or other AGI Agents. It enables the possibility of cooperative tasks and allows our robot to function as part of a larger system, fostering teamwork, and cooperative intelligence.

The transition to this proposed framework presents significant advantages. Firstly, it improves the scalability and adaptability of the system. Algorithms in the AI Agent module can be designed and deployed across different types of robots without needing to understand the intricacies of each specific hardware component. Secondly, the framework encourages modularity, making it easier to add, remove, or upgrade specific functions. Lastly, the introduction of the External Layer opens up new possibilities for human-robot interaction and cooperative intelligence, fostering the integration of the robot into larger, more complex systems.

\section{Conclusion}
In this work, we have proposed a novel control paradigm for autonomous robots, unifying the intellectual capabilities of large language models (LLMs) with instinctual functionalities. Our layered design approach inherently caters to the evolving demand for intelligent, safe, and adaptive robotic systems.

The burgeoning development of AI agents, particularly LLMs, has been a game-changer in the realm of robotics, offering immense benefits such as intuitive task comprehension, adept human-robot interaction, and complex decision-making. However, with these opportunities come inherent risks, especially regarding safety. By integrating an Instinct Layer in our paradigm, we ensure the prompt execution of survival-essential tasks and maintain an upper hand in safety matters.

Despite the promising blueprint of our new architecture, there are avenues left unexplored in this paper. Currently, we are planning to conduct experimental validations of our paradigm on mobile robot platforms and multi-axis robotic arms. This will further substantiate the enhanced intelligence and safety aspects brought by our design. Furthermore, the inter-layer communication, although not covered extensively in this paper, is a crucial area of future research. We envision a medium akin to Robot OS between the Decision and Instinct Layers to bolster system transparency, reduce design complexity, and increase scalability.

In addition, we barely scratched the surface of long/short term memory in this paper. A more profound exploration is planned for future studies, elucidating how AI Agents and Robot Instincts can learn and optimize strategies from their memory.

While our work indeed poses challenges and has room for refinement, the proposed architecture unarguably opens up new frontiers in the landscape of autonomous robotics. By bridging intelligence and instinct, we hope to harness the immense potential of AI and robotics and anticipate a future where robots can operate independently, intelligently, and safely in a wide range of environments.

%%%%%%%%%%%%%%%%%%%%%%%%%%%%%%%%%%%%%%%%%%%%%%%%%%%%%%%%%%%%%%%%%%%%%%%%%%%%%%%%
% \section*{APPENDIX}

% Appendixes should appear before the acknowledgment.

% \section*{ACKNOWLEDGMENT}
% We thank Hrstek Artificial Intelligence Company for their invaluable help and support during the completion of this article. We express our sincere gratitude to the dedicated team at Hrstek for their expertise and guidance.

%%%%%%%%%%%%%%%%%%%%%%%%%%%%%%%%%%%%%%%%%%%%%%%%%%%%%%%%%%%%%%%%%%%%%%%%%%%%%%%%

% References are important to the reader; therefore, each citation must be complete and correct. If at all possible, references should be commonly available publications.

{\small
\bibliographystyle{ieee_fullname}
\bibliography{reference}
}

\end{document}